\pdfoutput=1

\documentclass[11pt]{article}

\usepackage{acl}

\usepackage{times}
\usepackage{latexsym}
\usepackage{booktabs}

\usepackage[T1]{fontenc}

\usepackage{graphicx}

\usepackage[utf8]{inputenc}

\usepackage{microtype}

\title{A global analysis of metrics used for measuring performance in natural language processing}

\author{Kathrin Blagec \and {\bf Georg Dorffner} \and {\bf Milad Moradi} \and \\ {\bf Simon Ott} \and {\bf Matthias Samwald} \\
Institute of Artificial Intelligence; \\Center for Medical Statistics, Informatics, and Intelligent Systems; \\Medical University of Vienna, Vienna, Austria. \\\\}

\begin{document}

\maketitle
\begin{abstract}

Measuring the performance of natural language processing models is challenging. Traditionally used metrics, such as BLEU and ROUGE, originally devised for machine translation and summarization, have been shown to suffer from low correlation with human judgment and a lack of transferability to other tasks and languages. In the past 15 years, a wide range of alternative metrics have been proposed. However, it is unclear to what extent this has had an impact on NLP benchmarking efforts. Here we provide the first large-scale cross-sectional analysis of metrics used for measuring performance in natural language processing.
We curated, mapped and systematized more than 3500 machine learning model performance results from the open repository ‘Papers with Code’ to enable a global and comprehensive analysis. Our results suggest that the large majority of natural language processing metrics currently used have properties that may result in an inadequate reflection of a models’ performance. Furthermore, we found that ambiguities and inconsistencies in the reporting of metrics may lead to difficulties in interpreting and comparing model performances, impairing transparency and reproducibility in NLP research.

\end{abstract}

\section{Introduction}

Benchmarking, i.e., the process of measuring and comparing model performance on a specific task or set of tasks, is an important driver of progress in natural language processing (NLP). Benchmark datasets are conceptualized as fixed sets of data that are manually, semi-automatically or automatically generated to form a representative sample for these specific tasks to be solved by a model. A model’s performance on such a benchmark is then assessed based on a single or a small set of performance metrics. While this enables quick comparisons, it may entail the risk of conveying an incomplete picture of model performance since metrics inherently condense performance to a single number, omitting certain performance aspects completely or balancing trade-offs between different aspects (e.g. accuracy vs. fluency). Additionally the capacity of metrics to capture performance may differ strongly between tasks and languages.

Capturing model performance in a single metric is an inherently difficult task, and this is further aggravated in the NLP domain by the structural and semantic complexity of human language. Traditionally used NLP metrics such as BLEU or ROUGE, originally devised for machine translation and summarization, were shown to suffer from low correlation with human judgment and poor transferability to other tasks \citep{lin_2004_1, liu_2008, ng_2015, novikova_2017, chen_2019}. These fundamental problems are increasingly recognized by the NLP community—e.g., metric evaluation was even introduced as an independent task at the annual Machine Translation conference \citep{ma_2019}.

In the past 15 years, a wide variety of superior metrics for evaluating models on NLP tasks have been proposed, including task-agnostic, AI-based metrics such as BERTscore \citep{zhang_2019, peters_2018, clark_2019}. However, it is unknown to what extent this had an impact on metrics used in NLP research.

We aim to address this question by providing a global analysis of performance measures used in NLP benchmarking. Our contributions are three-fold: (1) We curated, mapped and systematized performance metrics covering more than 3500 performance results from the open repository ‘Papers with Code’ to enable a global and comprehensive analysis. (2) Based on this dataset, we provide a cross-sectional analysis of the prevalence of performance measures in the subset of natural language processing benchmarks. (3) We describe inconsistencies and ambiguities in the reporting and usage of metrics, which may lead to difficulties in interpreting and comparing model performances.

\section{Methods}

\subsection{Dataset}

Our analyses are based on data available from Papers with Code (PWC), a large, web-based open platform that collects Machine learning papers and summarizes evaluation results on benchmark datasets. PWC is built on automatically extracted data from arXiv submissions and manual crowd-sourced annotation.

The Intelligence Task Ontology (ITO) aims to provide a comprehensive map of artificial intelligence tasks using a richly structured hierarchy of processes, algorithms, data and performance metrics.\footnote{https://github.com/OpenBioLink/ITO} ITO is based on data from PWC and the EDAM ontology \footnote{http://edamontology.org/}. The development process of ITO is detailed in \citep{blagec_website_2021}. We built on ITO for further curation and on of a hierarchical mapping of the raw performance metric data from PWC.

\subsection{Hierarchical mapping and further curation of metric names}

The raw dataset exported from PWC contained a total number of 812 different strings representing metric names that appeared as distinct data property instances in ITO. These metric names were used by human annotators on the PWC platform to add results for a given model to the evaluation table of the relevant benchmark dataset’s leaderboard on PWC. This list of raw metrics in the PWC database was manually curated into a canonical hierarchy by our team. This entailed some complexities and required extensive manual curation which was conducted based on the mapping proceduce described below.

In many cases, the same metric was reported under multiple different synonyms and abbreviations. Furthermore, many results were reported in specialized sub-variants of established metrics. For each metric a canonical property denoting its general form (e.g., 'BLEU score') was created, and synonyms and sub-variants were mapped to it. For example, the reported performance metrics ‘BLEU-1’, ‘BLEU-2’ and ‘B-3’ were made sub-metrics of 'BLEU score'. Throughout the paper, we will refer to canonical properties and mapped metrics as ‘top-level metrics’ and ‘sub-metrics’, respectively.

In case a library that implemented a metric was used as the metric name (e.g., SacreBLEU, which is a reference implementation of the BLEU score available as a Python package), this property was made sub-metric of the more general metric name, in this case ‘BLEU score’.

271 entries from the original list could not be assigned a metric and were subsumed under a separate category ‘Undetermined’. After this extensive manual curation, the resulting list covered by our dataset could be reduced from 812 to 187 distinct performance metrics. Where possible, we used the respective preferred Wikipedia article titles as canonical names for the metrics. For an excerpt of the resulting property hierarchy, see Figure \ref{figure:propertyhierarchy} in Appendix \ref{appendix:a}.

\subsection{Grouping of top-level metrics}

Top-level metrics were further grouped into categories based on the task type they are usually applied to: Classification, Computer vision, Natural language processing, Regression, Game playing, Ranking, Clustering and ‘Other’. We limited our main analysis to the category ‘Natural language processing’, which only contains metrics that are specific to NLP, such as ROUGE, BLEU or METEOR. We provide additional statistics on general classification metrics, such as Accuracy or F1 score that are also often used in NLP benchmarks but are not specific to NLP tasks in Table \ref{table:classification} in Appendix \ref{appendix:b}.

\subsection{Analysis}

Analyses were performed based on the ITO release of 13.7.2020. Raw statistics were generated based on the ITO ontology using SPARQL queries and further processed and analyzed using Jupyter Notebooks and the Python ‘pandas’ library. Data, code and notebooks to generate these statistics are available on Github (see section ‘Data and code availability’).

\section{Results}

\subsection{Data basis}

\begin{table*}
\centering
\begin{tabular}{@{}lcl@{}}
\toprule
                                                & \multicolumn{1}{l}{Total dataset} & NLP subset \\ \midrule
Number of benchmark datasets                    & 2,298                             &        491    \\
Number of benchmark results                     & 32,209                            &        4,812    \\
Time span covered & 2000-2020                         &    2000-2020        \\ \bottomrule
\end{tabular}
\caption{General descriptives of the analyzed dataset (as of July 2020).}
\label{tab:datasetdescriptives}
\end{table*}

32,209 benchmark results across 2,298 distinct benchmark datasets reported in a total number of 3,867 papers were included in this analysis. Included papers consist of papers in the PWC database that were annotated with at least one performance metric as of July 2020. A single paper can thus contribute results to more than one benchmark and to one or more performance metrics.

The publication period of the analyzed papers covers twenty years, from 2000 until 2020, with the majority having been published in the past ten years (see Figure \ref{figure:publicationcount} in Appendix \ref{appendix:b}). 

The subset of NLP benchmark datasets considered in our analysis included 4,812 benchmark results across 491 benchmark datasets (see Table \ref{tab:datasetdescriptives}).

\subsection{Which performance metrics are most frequently reported in NLP benchmarking?}

Table \ref{tab:top10} lists the top 10 most frequently reported performance metrics. Considering submetrics, ROUGE-1, ROUGE-2 and ROUGE-L were the most commonly annotated ROUGE variants, and BLEU-4 and BLEU-1 were the most frequently annotated BLEU variants. For a large fraction of BLEU and ROUGE annotations, the subvariant was not specified in the annotation.

\begin{table*}
\centering
\begin{tabular}{@{}lcl@{}}
\toprule
Performance metric         & \multicolumn{1}{l}{Number of benchmark datasets} & Percent \\ \midrule
BLEU score                 & 300                                              & 61.1    \\
ROUGE metric               & 114                                              & 23.2    \\
Perplexity                 & 48                                               & 9.8     \\
METEOR                     & 39                                               & 7.9     \\
Word error rate            & 36                                               & 7.3     \\
Exact match                & 33                                               & 6.7     \\
CIDEr                      & 24                                               & 4.9     \\
Unlabeled attachment score & 18                                               & 3.7     \\
Labeled attachment score   & 15                                               & 3.1     \\
Bit per character          & 12                                               & 2.4     \\ \bottomrule
\end{tabular}
\caption{Top 10 reported NLP metrics and percent of NLP benchmark datasets (n=491) that use the respective metric. BLEU: Bilingual Evaluation Understudy, CIDEr: Consensus-based Image Description Evaluation, ROUGE: Recall-Oriented Understudy for Gisting Evaluation, METEOR: Metric for Evaluation of Translation with Explicit ORdering.}
\label{tab:top10}
\end{table*}

The BLEU score was used across a wide range of NLP benchmark tasks, such as machine translation, question answering, summarization and text generation. ROUGE metrics were mostly used for text generation, video captioning and summarization tasks while METEOR was mainly used for image and video captioning, text generation and question answering tasks.

\subsection{Are metrics reported together with other metrics or do they stand alone?}

The BLEU score was reported without any other metrics in 80.2\% of the cases, whereas the ROUGE metrics more often appeared together with other metrics and stood alone in only nine out of 24 occurrences. METEOR was, in all cases, reported together with at least one other metric. Figure \ref{figure:coocurrence} in Appendix \ref{appendix:b} shows the co-occurrence matrix for the top 10 most frequently used NLP-specific metrics. BLEU was most often reported together with the ROUGE metrics (n=12) and METEOR (n=12). ROUGE likewise frequently appeared together with METEOR (n=10). We additionally provide statistics on the number of distinct metrics per benchmark for the total dataset in Figure \ref{figure:distinctmetrics} in Appendix \ref{appendix:b}.

\subsection{Inconsistencies and ambiguities in the reporting of performance metrics}

During the mapping process it became evident that performance metrics are often reported in an inconsistent or ambiguous manner. One example for this are the ROUGE metrics, which have originally been proposed in different variants (e.g., ROUGE-1, ROUGE-L) but are often simply referred to as ‘ROUGE’. Furthermore, ROUGE metrics have originally been proposed in a ‘recall’ and ‘precision’ sub-variant, such as ‘ROUGE-1 precision’ and ‘ROUGE-1 recall’. Further, the harmonic mean between these two scores (ROUGE-1 F1 score) can be calculated. However, results are often reported as, e.g., 'ROUGE-1' without specifying the variant, which may lead to ambiguities when comparing results between different publications.

\section{Discussion}

NLP covers a wide range of different tasks and thus shows a large diversity of utilized metrics. We limited our analysis to more complex NLP tasks beyond simple classification, such as machine translation, question answering, and summarization. Metrics designed for these tasks generally aim to assess the similarity between a machine-generated text and a reference text or set of reference texts that are human-generated.

We found that, despite their known shortcomings, the BLEU score and ROUGE metrics continue to be the most frequently used metrics for such tasks.

Several weaknesses of BLEU have been pointed out by the research community, such as its sole focus on n-gram precision without considering recall and its reliance on exact n-gram matchings. Zhang et al. have discussed properties of the BLEU score and NIST, a variant of the BLEU score that gives more weight to rarer n-grams than to more frequent ones, and came to the conclusion that neither of the two metrics necessarily show high correlation with human judgments of machine translation quality \citep{doddington_website_2002, zhang_2004}.

\begin{figure*}
\centering
\includegraphics[width=6in]{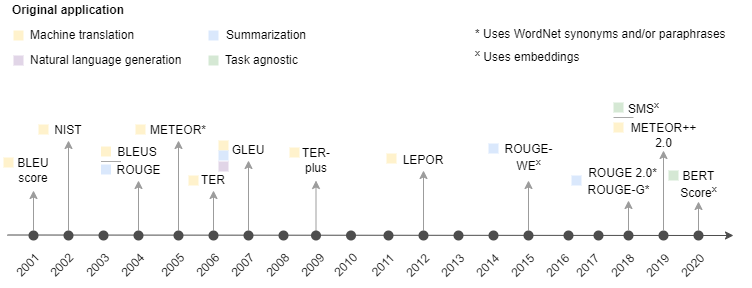}
\caption{Timeline of the introduction of NLP metrics and their original application. SMS: Sentence Mover's Similarity.}
\label{figure:timeline}
\end{figure*}

The ROUGE (Recall-Oriented Understudy for Gisting Evaluation) metrics family was the second most used NLP-specific metric in our dataset after the BLEU score. While originally proposed for summarization tasks, a subset of the ROUGE metrics (i.e., ROUGE-L, ROUGE-W and ROUGE-S) has also been shown to perform well in machine translation evaluation tasks \citep{lin_2004_1, och_2004}. However, the ROUGE metrics set has also been shown to not adequately cover multi-document summarization, tasks that rely on extensive paraphrasing, such as abstractive summarization, and extractive summarization of multi-logue text types (i.e., transcripts with many different speakers), such as meeting transcripts \citep{lin_2004_1, liu_2008, ng_2015}. Several new variants have been proposed in recent years, which make use of the incorporation of word embeddings (ROUGE-WE), graph-based approaches (ROUGE-G), or the extension with additional lexical features (ROUGE 2.0) \citep{ng_2015, shafieibavani_2018, ganesan_2018}. ROUGE-1, ROUGE-2 and ROUGE-L were the most common ROUGE metrics in our analyzed dataset, while newer proposed ROUGE variants were not represented.

METEOR (Metric for Evaluation of Translation with Explicit Ordering) was proposed in 2005 to address weaknesses of previous metrics \citep{banerjee_2005}. METEOR is an F-measure derived metric that has repeatedly been shown to yield higher correlation with human judgment across several tasks as compared to BLEU and NIST \citep{lavie_2004, graham_2015, chen_2019}. Matchings are scored based on their unigram precision, unigram recall (given higher weight than precision), and a comparison of the word ordering of the translation compared to the reference text. This is in contrast to the BLEU score, which does not take into account n-gram recall. Furthermore, while BLEU only considers exact word matches in its scoring, METEOR also takes into account words that are morphologically related or synonymous to each other by using stemming, lexical resources and a paraphrase table. Additionally, METEOR was designed to provide informative scores at sentence-level and not only at corpus-level. An adapted version of METEOR, called METEOR++ 2.0, was proposed in 2019 \citep{guo_2019}. This variant extends METEOR’s paraphrasing table with a large external paraphrase database and has been shown to correlate better with human judgement across many machine translation tasks. 

Compared to BLEU and ROUGE, METEOR was rarely used as a performance metric (8\%) across the NLP benchmark datasets included in our dataset.

The GLEU score was proposed as an evaluation metric for NLP applications, such as machine translation, summarization and natural language generation, in 2007 \citep{mutton_2007}. It is a Support Vector Machine-based metric that uses a combination of individual parser-derived metrics as features. GLEU aims to assess how well the generated text conforms to ‘normal’ use of human language, i.e., its ‘fluency’. This is in contrast to other commonly used metrics that focus on how well a generated text reflects a reference text or vice versa. GLEU was reported only in 1.8\% of NLP benchmark datasets.

Additional alternative metrics that have been proposed by the NLP research community but do not appear as performance metrics in the analyzed dataset include Translation error rate (TER), TER-Plus, “Length Penalty, Precision, n-gram Position difference Penalty and Recall” (LEPOR), Sentence Mover’s Similarity, and BERTScore. Figure \ref{figure:timeline} depicts the timeline of introduction of NLP metrics and their original application.

TER was proposed as a metric for evaluating machine translation quality. TER measures quality by the number of edits that are needed to change the machine-generated text into the reference text(s), with lower TER scores indicating higher translation quality \citep{makhoul_2006}. TER considers five edit operations to change the output into the reference text: Matches, insertions, deletions, substitutions and shifts. An adaptation of TER, TER-Plus, was proposed in 2009. Ter-Plus extends TER with three additional edit operations, i.e., stem matches, synonym matches and phrase substitution \citep{snover_2009}. TER-Plus was shown to have higher correlations with human judgements in machine translation tasks than BLEU, METEOR and TERp \citep{snover_2009}.
LEPOR and its variants hLEPOR and nLEPOR were proposed as a language-independent model that aims to address the issue that several previous metrics tend to perform worse on languages other than those it was originally designed for. It has been shown to yield higher correlations with human judgement than METEOR, BLEU, or TER \citep{han_2012}.

Sentence Mover’s Similarity (SMS) is a metric based on ELMo word embeddings and Earth mover’s distance, which measures the minimum cost of turning a set of machine generated sentences into a reference text’s sentences \citep{peters_2018, clark_2019}. It was proposed in 2019 and was shown to yield better results as compared to ROUGE-L in terms of correlation with human judgment in summarization tasks.

BERTScore was proposed as a task-agnostic performance metric in 2019 \citep{zhang_2019}. It computes the similarity of two sentences based on the sum of cosine similarities between their token’s contextual embeddings (BERT), and optionally weighs them by inverse document frequency scores \citep{devlin_2018}. BERTScore was shown to outperform established metrics, such as BLEU, METEOR and ROUGE-L in machine translation and image captioning tasks. It was also more robust than other metrics when applied to an adversarial paraphrase detection task. However, the authors also state that BERTScore’s configuration should be adapted to task-specific needs since no single configuration consistently outperforms all others across tasks.

Difficulties associated with automatic evaluation of machine generated texts include poor correlation with human judgement, language bias (i.e., the metric shows better correlation with human judgment for certain languages than others), and worse suitability for language generation tasks other than the one it was proposed for \citep{novikova_2017}. In fact, most NLP metrics have originally been conceptualized for a very specific application, such as BLEU and METEOR for machine translation, or ROUGE for the evaluation of machine generated text summaries, but have since then been introduced as metrics for several other NLP tasks, such as question-answering, where all three of the above mentioned scores are regularly used. Non-transferability to other tasks has recently been shown by Chen et al. who have compared several metrics (i.e., ROUGE-L, METEOR, BERTScore, BLEU-1, BLEU-4, Conditional BERTScore and Sentence Mover’s Similarity) for evaluating generative Question-Answering (QA) tasks based on three QA datasets. They recommend that from the evaluated metrics, METEOR should preferably be used and point out that metrics originally introduced for evaluating machine translation and summarization do not necessarily perform well in the evaluation of question answering tasks \citep{chen_2019}.

Many NLP metrics use very specific sets of features, such as specific word embeddings or linguistic elements, which may complicate comparability and replicability. To address the issue of replicability, reference open source implementations have been published for some metrics, such as, ROUGE, sentBleu-moses as part of the Moses toolkit and sacreBLEU \citep{lin_2004_1}.

In summary, we found that the large majority of metrics currently used to report NLP research results have properties that may result in an inadequate reflection of a models’ performance. While several alternative metrics that address problematic properties have been proposed, they are currently rarely used in NLP benchmarking. Our findings are in line with a recent, focused meta-analysis on machine translation conducted by Marie et al. who found that 82.1\% of papers report BLEU as the only performance metric despite its well-known shortcomings \citep{marie_2021}. Our analysis extends these findings by providing a global overview of metrics used in the entire NLP domain. 

\subsection{Recommendations for reporting performance results and future considerations}
In the following, we provide recommendations on the reporting of performance metrics and discuss potential future avenues for improving measuring performance using benchmarks in NLP.

\subsubsection{Increasing transparency and consistency in the reporting of performance metrics}
Performance metrics should be reported in a clear and unambiguous way to improve transparency, avoid misinterpretation and enable reproducibility.

\begin{itemize}
\item For performance metrics that have various sub-variants, it should be clearly stated which variant is reported (e.g., ROUGE-1 F1 score instead of ROUGE-1). If multiple metrics are averaged, it should be stated what kind of mean is used (e.g., arithmetic mean, geometric mean, harmonic mean) if this is not clear from the definition of the metric itself (e.g., F1 score). 
\item If a metric is used that allows for adaptations, such as weighting, these should be explicitly stated and be marked clearly in the result tables. Ideally, when using abbreviations, the variant should be included in the abbreviation or e.g., marked by a subscript.
\item To increase transparency and allow reproducibility, the formula for calculating the metric should be included in the manuscript or in the Appendix.
\item For more complex metrics, if available, a reference implementation should be used and cited. If such a reference implementation is not available, or a custom implementation or adaptation is used, the code should be made available.
\end{itemize}

In the future, a taxonomic hierarchy of performance metrics that captures definitions, systematizes metrics together with all existing variants and lists recommended applications based on comparative evaluation studies. In this work, we have created a starting point for creating such a taxonomy using a bottom-up approach as part of ITO \citep{blagec_website_2021}. 

\subsubsection{Maximizing the informative value in the reporting of performance results}

Developing metrics for NLP tasks is an ongoing research area, new metrics outperforming previous ones are proposed on a regular basis, and suitability is strongly task- and dataset-dependent, therefore general advice on which metric to use cannot be given.

Instead, it should be critically evaluated whether a metric is suitable for a given dataset, task or language, especially if the metric was originally proposed for a different application. Comparative evaluation studies, such as in \citep{chen_2019} can provide an indication for the suitability.

If a metric is used that has been shown to have limited informative value (in general, or in specific use cases) and no alternative is available, the limitations and their relevance for the task and/or dataset should be discussed.

If more than one suitable metric is available, consider reporting all of them, especially if there is a discrepancy in performance results. 

Even if a benchmark is historically evaluated based on a certain metric, consider additionally reporting newer proposed metrics if they are suitable and have been evaluated to be useful for the task.

\subsection{Future considerations on performance metrics in the context of benchmarking}
Comparative evaluation studies investigating performance metrics, their properties and their correlation across multiple tasks, datasets and languages could help to better understand metrics and their suitability for different applications. While studies focusing on a small set of metrics exist, such as in \citep{chen_2019}, larger studies are, to the best of our knowledge, yet to be undertaken.

Recent work introduced the notions of dynamic benchmarks that allow users to weigh different performance metrics of interest. An example of this is ‘Dynascore’ which allows customizable aggregation of performance across different aspects including non-traditionally assessed performance dimensions, such as memory, robustness, and “fairness \citep{Ma_Ethayarajh_Thrush_Jain_Wu_Jia_Potts_Williams_Kiela_2021}.
Further, bidimensional leaderboards based on linear ensembles of metrics have been proposed \citep{gehrmann_2021, ruder2021benchmarking, kasai_2021}. These approaches could further improve the practical utility of benchmark results.

\subsection{Limitations}
Our analyses are based on ITO v0.21 which encompasses data until mid 2020. To ensure that our results are still relevant given the fast pace of research, we checked whether considering data from the recently released ITO v1.01 which includes data until mid 2021 leads to any significant time-dependent changes of our results \footnote{Data curation in ITO v1.01 is still incomplete. Therefore, results are based on the fully curated ITO v0.21.}. Including this more recent data did, however, not alter the described usage patterns of NLP metrics.

The results presented in this paper are based on a large set of machine learning papers available from the PWC database, which is the largest annotated dataset of benchmark results currently available. The database comprises both preprints of papers published on arXiv and papers published in peer-reviewed journals. While it could be argued that arXiv preprints are not representative of scientific journal articles, it has recently been shown that a large fraction of arXiv preprints (77\%) are subsequently published in peer-reviewed venues \citep{lin_2020}.

\section{Conclusions}

The reporting of metrics was partly inconsistent and partly unspecific, which may lead to ambiguities when comparing model performances, thus negatively impacting the transparency and reproducibility of NLP research. Large comparative evaluation studies of different NLP-specific metrics across multiple benchmarking tasks are needed.

\section*{Data and code availability}

The OWL (Web Ontology Language) file of the ITO model is made available on Github \footnote{https://github.com/OpenBioLink/ITO} and BioPortal \footnote{https://bioportal.bioontology.org/ontologies/ITO}. The ontology file is distributed under a CC-BY-SA license. ITO includes data from the Papers With Code project \footnote{https://paperswithcode.com/}. Papers With Code is licensed under the CC-BY-SA license. Data from Papers With Code are partially altered (manual curation to improve ontological structure and data quality). ITO includes data from the EDAM ontology. The EDAM ontology is licensed under a CC-BY-SA license.

Notebooks containing the queries and code for data analysis are also accessible via GitHub.

\section*{Acknowledgements}

We thank the team from 'Papers With Code' for making their database available and all annotators who contributed to it.

\bibliography{anthology,custom}
\bibliographystyle{acl_natbib}

\appendix
\counterwithin{figure}{section}
\counterwithin{table}{section}

\section{Appendix: Performance metric property hierarchy}
\label{appendix:a}

\begin{figure*}
\centering
\includegraphics[width=6in]{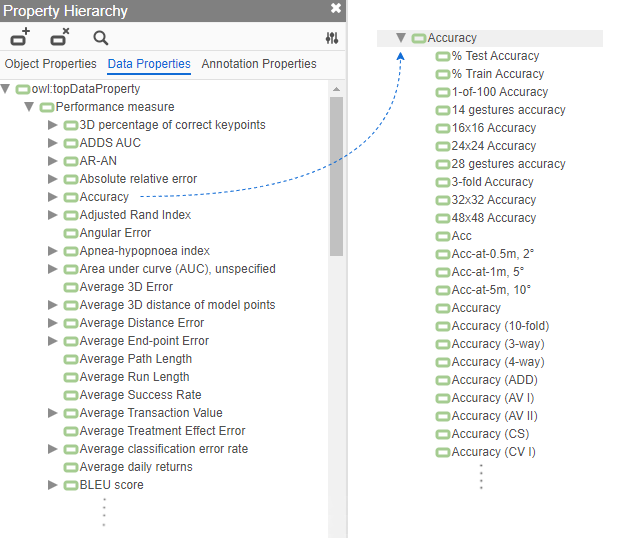}
\caption{Property hierarchy after manual curation of the raw list of metrics. The left side of the image shows an excerpt of the list of top-level performance metrics; the right side shows an excerpt of the list of submetrics for the top-level metric ‘Accuracy’.}
\label{figure:propertyhierarchy}
\end{figure*}

\section{Appendix: Additional statistics on the dataset}
\label{appendix:b}

\begin{figure*}
\centering
\includegraphics[width=6.3in]{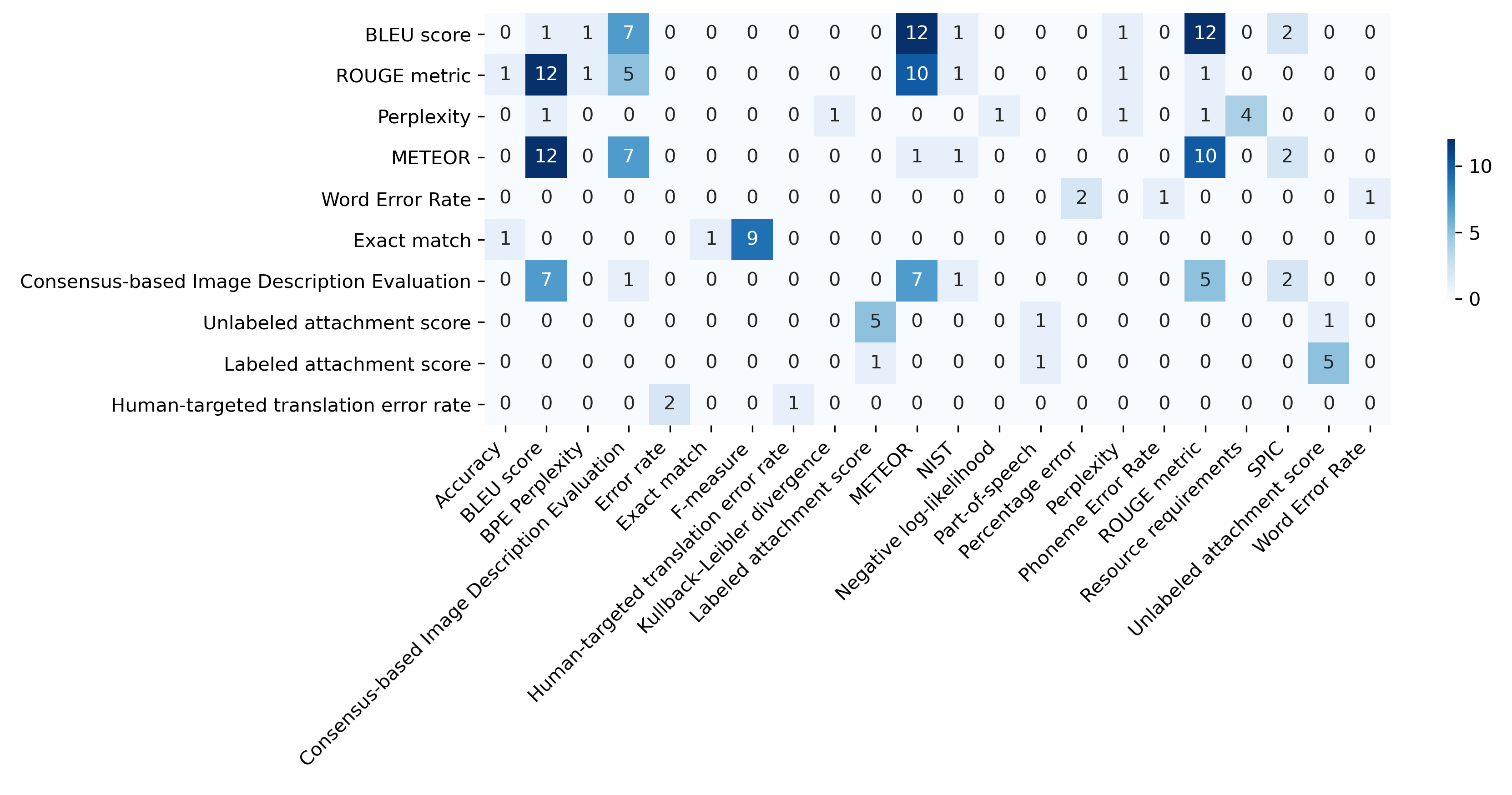}
\caption{Co-occurrence matrix for the top 10 most frequently used NLP metrics (y-axis). Only metrics that were reported at least one time together with either one of the selected metrics are shown (x-axis).}
\label{figure:coocurrence}
\end{figure*}

\begin{figure*}
\centering
\includegraphics[width=5in]{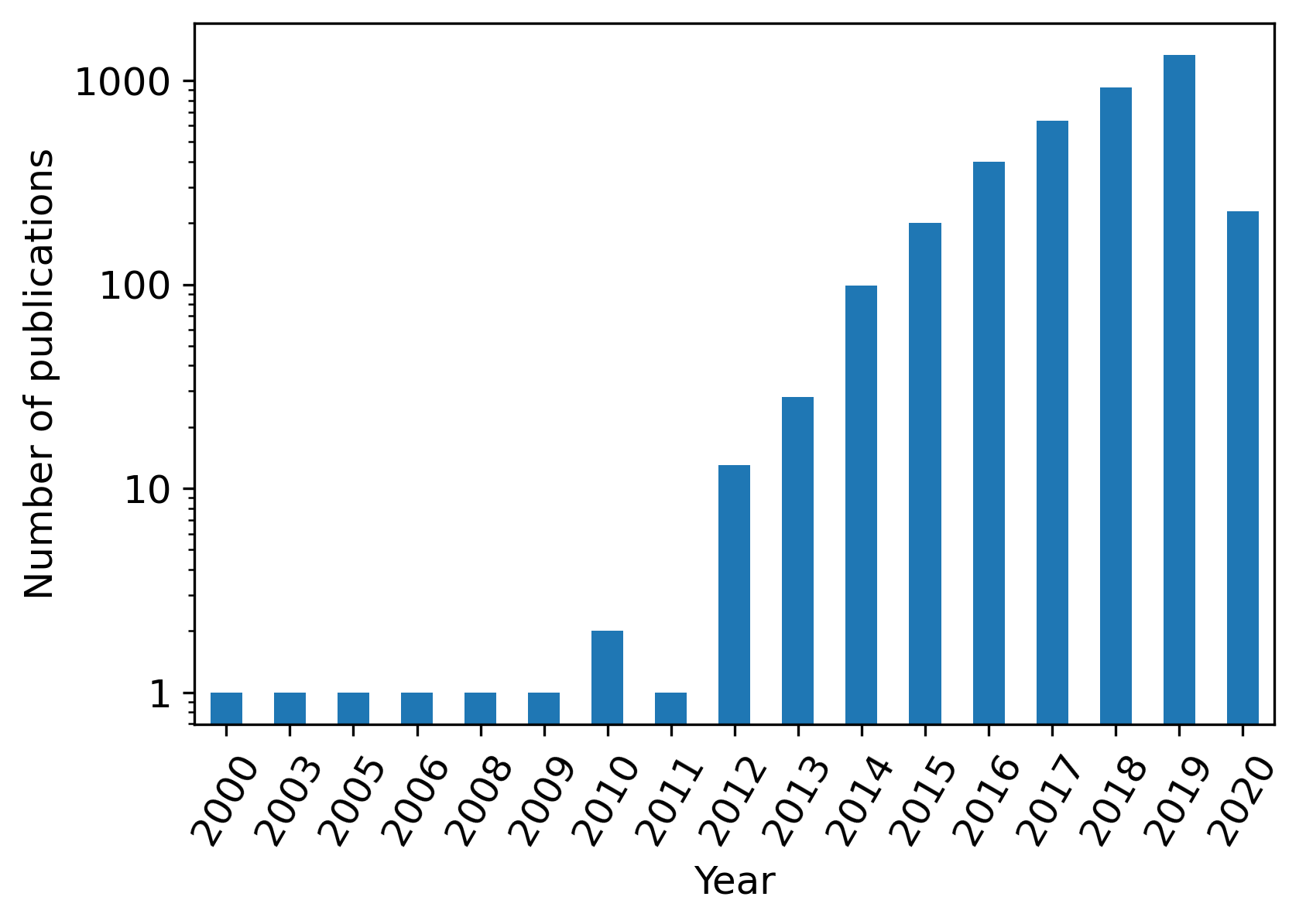}
\caption{ Number of publications covered by the total dataset per year. The y-axis is scaled logarithmically. }
\label{figure:publicationcount}
\end{figure*}

\begin{table*}
\centering
\begin{tabular}{lcl}
\hline
Performance metric & \multicolumn{1}{l}{Number of benchmark datasets} & Percent \\ \hline
Accuracy           & 871                                              & 37.9    \\
F-measure          & 393                                              & 17.1    \\
Precision          & 374                                              & 16.3    \\
R@k             & 143                                              & 6.2     \\
AUC                & 123                                              & 5.4     \\
IoU                & 115                                              & 5.0     \\
Recall             & 79                                               & 3.4     \\
Hits@k          & 69                                               & 3.0     \\
P@k             & 33                                               & 1.4     \\
Error rate         & 30                                               & 1.3     \\ \hline
\end{tabular}
\caption{Top 10 reported simple classification metrics and percent of benchmark datasets that use the respective metric. R@k: Recall at k, AUC: Area under the curve, IoU: Intersection over union, P@k: Precision at k. AUC contains both ROC-AUC and PR-AUC.}
\label{table:classification}
\end{table*}

\begin{figure*}
\centering
\includegraphics[width=5in]{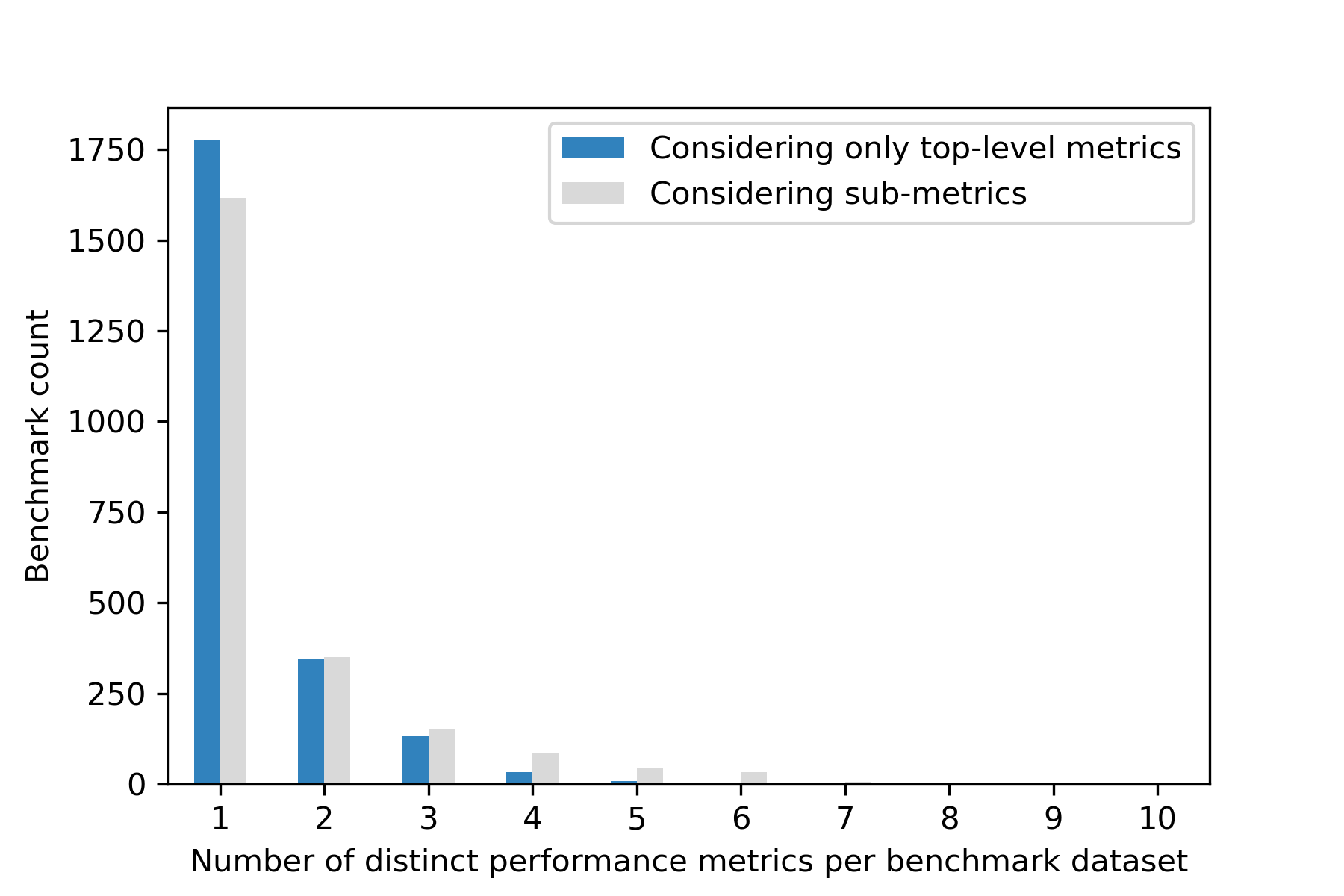}
\caption{ Count of distinct metrics per benchmark dataset when considering only top-level metrics as distinct metrics (blue bars), and when considering sub-metrics as distinct metrics (grey bars). Median number of distinct metrics per benchmark: 1. Data is shown for the complete dataset (n=2,298).}
\label{figure:distinctmetrics}
\end{figure*}

\end{document}